\documentclass[letterpaper, 10 pt, conference]{ieeeconf} % Comment this line out if you need a4paper

\IEEEoverridecommandlockouts % This command is only needed if 
\usepackage[table,xcdraw]{xcolor}
\usepackage[pdftex]{graphicx}
\usepackage{amssymb}
\usepackage{array}
\usepackage {booktabs}
\usepackage{url}
\newcolumntype{M}[1]{>{\centering\arraybackslash}m{#1}}
\newcolumntype{L}[1]{>{\arraybackslash}m{#1}}
\newcolumntype{C}{@{\extracolsep{3cm}}c@{\extracolsep{0pt}}}

\usepackage{color}

\newcommand{\cRyo}[1]{{\color{black}{#1}}} % Ryo Nakamura

\newcommand{\cNishiyama}[1]{{\color{black}{#1}}}

\title{\LARGE \bf
Traffic Incident Database with Multiple Labels Including \\ Various Perspective Environmental Information
}

\author{Shota Nishiyama*$^{1}$, Takuma Saito*$^{2}$, Ryo Nakamura$^{3,4}$,Go Ohtani$^{3,5}$\\
    Hirokatsu Kataoka$^{3}$ and Kensho Hara$^{3}$% <-this % stops a space
\thanks{*indicates equal contribution.}
\thanks{$^{1}$ Aichi Institute of Technology Graduate School of Business Administration and Computer Science, Toyota, 470-0356, Japan {\tt\small b21723bb@aitech.ac.jp}}%
\thanks{$^{2}$Tokyo Denki University, Tokyo, Japan, {\tt\small saito.t@is.fr.dendai.ac.jp}}%
\thanks{$^{3}$National Institute of Advanced Industrial Science and Technology (AIST), Ibaraki, Japan}%
\thanks{$^{4}$Fukuoka University, Fukuoka, Japan}%
\thanks{$^{4}$Keio University, Kanagawa, Japan}%
\\
\url{https://github.com/nissy-shota/V-TIDB}
}

\begin{document}

\maketitle
\thispagestyle{empty}
\pagestyle{empty}

\begin{abstract}
Traffic accident recognition is essential in developing automated driving and Advanced Driving Assistant System technologies.
A large dataset of annotated traffic accidents is necessary to improve the accuracy of traffic accident recognition using deep learning models.
Conventional traffic accident datasets provide annotations on the presence or absence of traffic accidents and other teacher labels, improving traffic accident recognition performance. However, the labels annotated in conventional datasets need to be more comprehensive to describe traffic accidents in detail.
Therefore, we propose V-TIDB, a large-scale traffic accident recognition dataset annotated with various environmental information as multi-labels. Our proposed dataset aims to improve the performance of traffic accident recognition by annotating ten types of environmental information as teacher labels in addition to the presence or absence of traffic accidents. V-TIDB is constructed by collecting many videos from the Internet and annotating them with appropriate environmental information.
In our experiments, we compare the performance of traffic accident recognition when only labels related to the presence or absence of traffic accidents are trained and when environmental information is added as a multi-label. In the second experiment, we compare the performance of the training with only ``contact level,'' which represents the severity of the traffic accident, and the performance with environmental information added as a multi-label.
The results showed that 6 out of 10 environmental information labels improved the performance of recognizing the presence or absence of traffic accidents. In the experiment on the degree of recognition of traffic accidents, the performance of recognition of car wrecks and contacts was improved for all environmental information. These experiments show that V-TIDB can be used to learn traffic accident recognition models that take environmental information into account in detail and can be used for appropriate traffic accident analysis.

\end{abstract}

\begin{figure}[t]
\centering
\includegraphics[width=8.75cm]{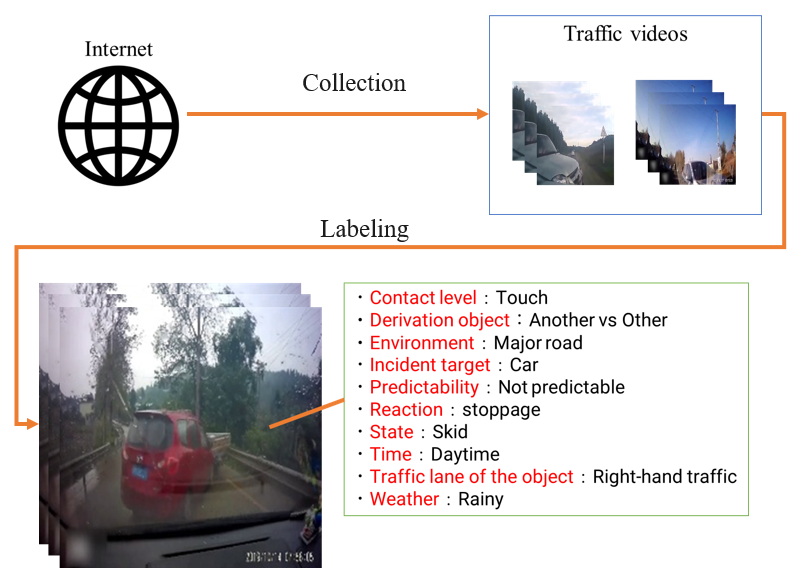}
\caption{Construction of a dataset labeled with various environmental information. The dataset was constructed by collecting a large number of traffic videos from the Internet and assigning labels indicating the presence or absence of traffic accidents and ten types of environmental information. }
 
\label{fig:abst}
\vspace{-10pt}
\end{figure}

\section{INTRODUCTION}

% 交通事故の認識は事前に
% 自動運転やADASの性能を高めることは，xxxという理由で重要である．
% その中でも特に運転による事故や
% 自動運転やADASの研究は、都市の交通シーンの動画データセットを通じて行われている。

\cRyo{With the development of robot technology, Advanced Driving Assistant Systems (ADAS) and automated driving are becoming more sophisticated. Studies of urban traffic scenes have contributed to the broadening of ADAS and automated driving \cite{cordts2016cityscapes}. The Honda Research Institute Driving Dataset (HDD) focuses on driving scene understanding and analyzes the interaction between humans and traffic scenes by detecting traffic participants and analyzing scenes as corresponding semantic categories \cite{Honda}. KITTI claims that autonomous driving systems rely on multiple sensors and environmental maps to provide video of a wide range of areas, including the periphery of medium-sized cities, rural areas, and highways \cite{KITII}. Berkeley DeepDrive Video (BDDV) aims to learn generic motion models to learn driving models and policies \cite{BDDV}.

Despite these advancements, 1.3 million people die in traffic accidents every year \cite{WHO}. Therefore, it is imperative to reduce the number of traffic accidents through the prediction and recognition of accidents by ADAS and automated driving systems. Additionally, even if automated vehicles and ADAS do not cause accidents, surrounding vehicles may cause accidents. Thus, it is expected that these systems must avoid traffic accidents by considering surrounding traffic participants (non-automated drivers, pedestrians, etc.). Avoiding traffic accidents is one of the most critical issues in automated driving and ADAS. Deep learning models are expected to improve the accuracy of traffic accident prediction and recognition for these systems.

Annotated traffic accident videos are necessary for training deep-learning models. However, the number of annotated traffic accident videos needs to be increased. The datasets proposed so far for traffic accident recognition have been annotated with information on the presence or absence of the accident, the region where the accident occurred, and the time of the accident. Some datasets provide time information of the moment when a traffic accident occurs. Other annotated information such as "Predictability," "Reaction," and "Traffic Lane" labels are also essential for traffic accident recognition.

Figure \ref{fig:abst} represents a sample traffic accident in one of our datasets. In this traffic accident, a car in the left lane "touched" an opposing car by skidding into the right lane. The cause of the skid was considered to be that the major road was wet due to rainy weather. Environmental information such as weather conditions and roads is essential in traffic accident recognition. However, deep learning models can be misleading if the surrounding environment information given in the video is insufficient. Thus, if the dataset does not contain annotations that explain environmental information, a model that cannot take environmental information into account is trained. The learned model will need help reflecting the influence of environmental information on the prediction and recognition of traffic accidents. In addition, the evaluation of the model may need to be revised when environmental information is not included.}

\cRyo{In this study, we propose V-TIDB, a dataset that includes ten types of environmental information in addition to the presence or absence of accident information in traffic accident videos. The ten types of environmental information in V-TIDB can be broadly divided into three elements: information around the accident, the accident itself, and the observer's point of view. The information about the accident includes local labels such as weather and time of day. The accident itself includes labels such as the vehicles involved and the degree of damage. The observer's point of view includes labels such as reactions and predictability. We report the construction of a new large-scale dataset, V-TIDB, a multi-label annotated dataset of ten types of detailed environmental information consisting of these three elements. We also provide a benchmark for traffic accident recognition on the V-TIDB dataset annotated with detailed environmental labels.}

In summary, our contributions are as follows:
\begin{itemize}
 \item We report that our multi-labeling of detailed environments improves the recognition performance of traffic accidents.
 \item  We propose a larger dataset than the previous dashcam traffic accident dataset.
 \item We report on constructing a dataset that includes observers' information to reveal more video information than conventional traffic accident datasets.
 \cNishiyama{
 \item We will publish links and annotations to the videos included in the V-TIDB.
 }
\end{itemize}

% \begin{table*}[t]
% % \begin{table}[h]
% \centering
% \caption{Comparisons of traffic incident datasets.
%     % Our dataset include a much larger number of videos and labels compared with conventional datasets.
% }
% % \caption{Videos from six datasets in the presence of 12 labels. This table shows that that the number of videos and labels in our dataset exceed those provided in conventional datasets.}
% \label{dataset_labels}

% \begin{tabular}{|l|c|c|c|c|c|c|c|}
% \hline

% Dataset & SA~\cite{SA} & A3D~\cite{A3D} & DADA~\cite{DADA} & DoTA~\cite{DoTA} & CCD~\cite{CCD} & NIDB~\cite{NIDB} & Ours \\ \hline \hline
% % Dataset & SA & A3D & DADA-2000 & DoTA & CCD & NIDB & Ours \\ \hline \hline
% \#Videos & 620 & 1500 & 2000 & 4677 & 4500 & 6200 & \textbf{9062} \\ \hline \hline
% Incident target & & & & \checkmark & & \checkmark & \checkmark \\ \hline
% Contact level & & & & & \checkmark & & \checkmark \\ \hline
% Derivation obj. & & \checkmark & & & & & \checkmark \\ \hline
% Environment & & & \checkmark & & & \checkmark & \checkmark \\ \hline
% Predictability & & & & & & & \checkmark \\ \hline
% Reaction & & & & & & & \checkmark \\ \hline
% State & & & & & \checkmark & & \checkmark \\ \hline
% Time & & & \checkmark & & \checkmark & \checkmark & \checkmark \\ \hline
% Traffic lane & & & & & & & \checkmark \\ \hline
% Weather & & & \checkmark & & \checkmark & & \checkmark \\ \hline
% Temporal & \checkmark & \checkmark & \checkmark & \checkmark & \checkmark & \checkmark & \\ \hline
% Spatial & & & \checkmark & \checkmark & \checkmark & \checkmark & \\ \hline
% \end{tabular}
% \
% \end{table*}

\begin{table*}[]
\caption{Comparisons of traffic incident datasets. Our data set is larger in variety than the previous data set.}
\centering
\label{tab:compare-dataset}
% \scalebox{0.9}{
\begin{tabular}{|l|wc{0.7cm}|wc{0.75cm}|wc{0.75cm}|wc{0.85cm}|wc{1.25cm}|wc{1.25cm}|wc{0.65cm}|wc{0.6cm}|wc{0.6cm}|wc{1.0cm}|wc{0.7cm}|wc{0.8cm}|wc{0.6cm}|}
\hline
Dataset      & \#Videos & \begin{tabular}[c]{@{}c@{}}Incident \\ target\end{tabular} & \begin{tabular}[c]{@{}c@{}}Contact \\ level\end{tabular} & \begin{tabular}[c]{@{}c@{}}Derivation \\ object\end{tabular} & Environment & Predictability & Reaction & State & Time & Traffic lane & Weather & Temporal & Spatial \\ \hline
SA {[}2{]}   & 620      &                                                            &                                                          &                                                              &             &                &          &       &      &              &         & \checkmark       &         \\ \hline
A3D {[}3{]}  & 1500     &                                                            &                                                          & \checkmark                                                           &             &                &          &       &      &              &         & \checkmark       &         \\ \hline
DADA {[}4{]} & 2000     &                                                            &                                                          &                                                              & \checkmark          &                &          &       & \checkmark   &              & \checkmark      & \checkmark       & \checkmark      \\ \hline
DoTA {[}6{]} & 4677     & \checkmark                                                         &                                                          &                                                              &             &                &          &       &      &              &         & \checkmark       & \checkmark      \\ \hline
CCD {[}5{]}  & 4500     &                                                            & \checkmark                                                       &                                                              &             &                &          & \checkmark    & \checkmark   &              & \checkmark      & \checkmark       & \checkmark      \\ \hline
NIDB {[}7{]} & 6200     & \checkmark                                                         &                                                          &                                                              & \checkmark          &                &          &       & \checkmark   &              &         & \checkmark       & \checkmark      \\ \hline
\textbf{Ours}        & 9062     & \checkmark                                                         & \checkmark                                                       & \checkmark                                                           & \checkmark          & \checkmark             & \checkmark       & \checkmark    & \checkmark   & \checkmark           & \checkmark      &          &         \\ \hline
\end{tabular}
% }
\vspace{-15pt}
\end{table*}

\section{Related work}
%シーン情報からの交通認識データセット
%動画像を用いた交通認識
\subsection{Traffic accident datasets}

There are two main types of existing traffic accident datasets: those captured by surveillance cameras and those captured by dashcams.

Surveillance videos provide a global view of multi-vehicle accidents but do not capture subjective factors that contribute to the accident. Additionally, it is challenging to communicate the results of traffic accident predictions to drivers using Time to Accident (TTA). A typical example of a surveillance camera dataset is the Traffic Accidents Dataset (TAD) \cite{TAD}, which primarily focuses on predicting traffic accidents on highways and includes weather and accident type labels.

Dashcam videos provide a driver's perspective and make it easy for drivers to understand the situation and predictions. The Street Accident (SA) dataset \cite{SA} is captured by dashcams and is used to predict accidents and detect participants who may have contributed to the accident. Another dashcam dataset, Car Crash Dataset (CCD) \cite{CCD}, includes weather labels in addition to time information, providing helpful information for predicting traffic accidents.

The Dataset of Object Detection in Aerial Images Detection of Traffic Anomaly (DoTA) \cite{DoTA} assumes that human attention deficits are a factor in traffic accidents and includes 4677 videos. Anomaly detection is used to detect traffic accidents, but it may be limited to binary classification of normal and abnormal. The Driver Attention Prediction in Driving Accidents (DADA-2000) dataset \cite{SegDADA} \cite{AugDADA} includes 2000 videos and has been extended to segmentation tasks to improve annotation quality.

The Near-miss Incident DataBase (NIDB) \cite{NIDB} includes 6300 videos and provides information on objects and environments to support the detection of near-miss incidents.

However, one challenge of these datasets is the limited amount of data. Therefore, some studies simulate traffic accident videos to predict and recognize traffic accidents. Examples of simulated datasets include GTA-Crash and Prescan \cite{GTACrash} \cite{Prescan}.

We propose a new dataset, V-TIDB, which consists of 9062 videos, including 4088 videos of traffic accidents. Our large traffic accident dataset addresses the challenges of anomaly detection and simulation datasets. It can support deep learning models to gain deeper insights than anomaly detection. Furthermore, the large amount of annotated environmental information greatly aids in exploring the perceptions and contributing factors of traffic accidents.

% 我々が提案するV-TIDBデータセットは、9062本の動画からなる大規模な交通事故データセットであり、V-TIDBには4088本の交通事故を含む動画が含まれている。私たちの大規模な交通事故データセットは、異常検出のためのデータセットとシミュレーションデータセットの課題を解決しています。私たちのデータセットは、異常検知よりも深い洞察を得るためのディープラーニングモデルをサポートすることができます。さらに、大量のアノテーションされた環境情報は、交通事故の認識や要因の探索に大いに役立ちます。

\subsection{Recognition task for video data using environmental information
}
Action recognition has been actively studied as a recognition task using video data.~\cite{Kinetics,UCFCrim,3DResnet}%[いくつかのデータセットを引用する]
It is known that using environmental information improves the robustness of action recognition, represented by video data recognition tasks.~\cite{electronics10192380,Multitask_Learning_action}
Recently, a dataset called Large Scale Holistic Video Understanding (HVU), which labels action labels and various environmental information, has been proposed\cite{HVU}.
\cRyo{The HVU dataset has been shown to improve the robustness of action classification when learning classifications based on labels of various environmental information.}
% Our paper can be positioned as an HVU dataset for traffic accident recognition systems.
% 行動認識は、ビデオデータを用いた認識タスクとして盛んに研究されている。~cite{Kinetics,UCFCrim,3DResnet}%[いくつかのデータセットを引用する]。
% ビデオデータ認識タスクに代表される行動認識において、環境情報を用いることで頑健性が向上することはよく知られている。
% 近年，行動ラベルと様々な環境情報をラベル付けしたLarge Scale Holistic Video Understanding (HVU) と呼ばれるデータセットが提案された．
% HVUデータセットでは、様々な環境情報のラベルに基づいて分類を学習する際に、行動分類の頑健性を向上させることも可能である。本論文は，交通事故認識システムのためのHVUデータセットとして位置づけることができる．

\subsection{Differences between conventional traffic accident recognition datasets}

The table\ref{tab:compare-dataset}  shows a comparison between our proposed dataset and the above dataset. Note that ``spatial" defines the bounding box of accident objects and is used in the accident detection task. At the same time, ``temporal" indicates the time of the accident and is used to learn the time before and after the accident. Here, our dataset has a much larger number of labels and videos. The labels assigned in previous studies tend to annotate the accident target and related environmental factors. Still, some labels annotate the ``reaction" or ``response" of the accident target, ``Was it an accident?" and ``Should I stop or swerve to prevent the accident?" and labels that answer subjective questions such as ``Was it an accident?" and ``Should I stop or swerve to prevent an accident?" are less common. In contrast, the labels added to the dataset proposed in this study describe the accident situation in more detail, allowing the evaluator to perform accident recognition activities more effectively.

% Tableref{tab:compare-dataset}は、我々の提案するデータセットと上記のデータセットの比較を示している。空間的 "は事故オブジェクトのバウンディングボックスを定義し、事故検出タスクで使用されることに注意。同時に、"temporal "は事故の時刻を示し、事故の前後の時刻を学習するために使用される。ここで、我々のデータセットでは、ラベルの数と動画の数が格段に多い。先行研究で割り当てられたラベルは、事故対象や関連する環境要因を注釈する傾向がある。それでも、事故対象の「反応」や「応答」を注釈するラベルもあり、「事故だったのか」「事故を防ぐために止まるべきか、ハンドルを切るべきか」といった主観的な問いに答えるラベルはあまり見かけません。これに対し、本研究で提案するデータセットに追加されたラベルは、事故状況をより詳細に記述しており、評価者はより効果的に事故認識活動を行うことができます。

\section{Various-perspective Traffic Incident Database \cRyo{V-TIDB}}
\cRyo{In this section, we introduce the V-TIDB (Various-perspective Traffic Incident Database), a proposed dataset for large-scale traffic recognition. In particular, we explain the two steps of creating a V-TIDB: ``collecting videos" and ``defining labels for videos". And finally, we explain the statistics of V-TIDB.}
% In this section, we describe our proposed dataset V-TIDB (Various-perspective Traffic Incident Database) for large-scale traffic recognition V-TIDB consists of more than 9000 videos collected from YouTube. V-TIDB also has 11 types of environmental information as multi-labels, which can be used to identify the risk and causes of traffic accidents.
% 本節では、大規模交通認識のための提案データセット V-TIDB (Various-perspective Traffic Incident Database) について説明する．
% 特に我々は，V-TIDBを作成手順である，"動画の収集"についてと"動画に付与するラベルの定義"について説明を行う．そして最後にV-TIDBの統計量を紹介する．

  % V-TIDB は YouTube から収集した 9000 以上の動画から構成されている。

% また、V-TIDBは11種類の環境情報をマルチラベルとして持っており、交通事故のリスクや原因の特定に利用することが可能である。

% 動画収集について
% 動画に付与するラベルについての定義
% データセットの統計値について紹介する．

\begin{figure*}[t]
\centering

\includegraphics[width=0.98\linewidth]{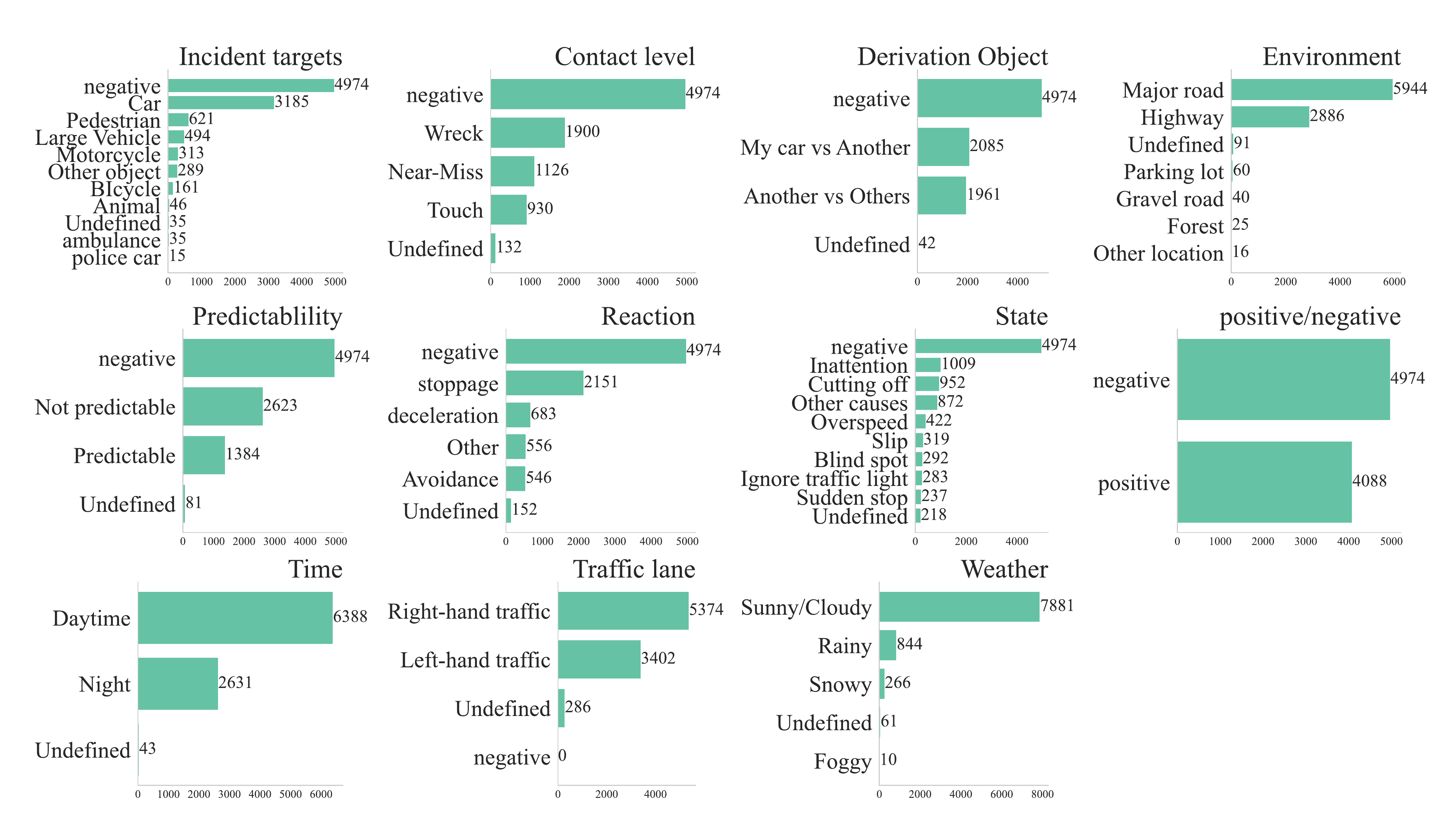}
\caption{The histogram shows the parent category name in the title and the child category name on the vertical axis. The horizontal axis is the number of child categories in V-TIDB.
}
\label{fig:statistic_vtidb}
\vspace{-15pt}
\end{figure*}

\subsection{Video Collection and Preprocessing}
% 我々は，交通事故を含むダッシュカム動画を集めるために，動画公開サイトであるYouTubeから動画の収集を行った．
% 収集した動画は，9000以上あり，これらは，「交通事故」「ヒヤリハッと」という検索ワードに基づき収集されました．
% それらの動画は，10秒以上ある場合，10秒に収まるように設定した．
% 交通事故を含む動画については，10秒の設定内に交通事故が含まれるように設定する．

% 動画の収集にYouTubeを用いる利点は，1)人力で収集するより比較的多くの動画を収集できる点と2)データセット公開時のプライバシー問題や倫理問題を回避できる点にある．手動で集めるより比較的低コストにかつ大量に動画を収集することを可能とする．
% 特にプライバシー問題と倫理問題については，交通事故認識データセットは交通事故を含む動画なので，倫理的に動画をオフィシャルに公開することを咎めらる可能性がある．そのため，動画を直接配布するのは困難であり，多くの交通事故データセットは公開されていない．動画収集にYotubeを用いる場合は，idを配布することによって，データセットを利用したいユーザーが自己責任での動画の収集を可能とするため，交通事故認識データセットにおいて相性が良い．
\cRyo{
We collected dashcam videos, including traffic accidents, from YouTube, a video open-source site. The videos are more than 9,000, and they were collected based on the search terms ``traffic accident" and ``near-miss. Those videos were set to fit into ten seconds if they were longer than ten seconds.
For videos that include traffic accidents, set the video to include traffic accidents within the ten-second setting. 

The advantages of using YouTube for video collection are 1) the ability to collect a relatively large number of videos compared to the human collection and 2) the ability to avoid privacy and ethical issues when releasing the dataset. It is possible to collect many videos at a relatively low cost compared to the human collection. In particular, regarding privacy and ethical issues, the Traffic Accident Recognition Dataset may be ethically censured from officially releasing the videos since the videos contain traffic accidents. Therefore, it is difficult to distribute videos directly, so many of the proposed traffic accident datasets are not directly distributed.
}
% The more than 9,000 videos included in V-TIDB were collected by more than ten annotators who searched for "traffic accidents" and "near-misses" on YouTube and collected only dashcam videos. Using a proprietary GUI(graphical user interface) tool, annotators selected appropriate annotations from 11 predefined parent labels and child classes. For videos longer than 10 seconds, the start time was set to 10 seconds before the end of the video. Only the time, weather, and traffic lane labels were assigned for videos that did not contain traffic incidents.
% V-TIDBに収録された9000本以上の動画は、10名以上のアノテーターがYouTubeで「交通事故」「ヒヤリハット」を検索し、ダッシュカム動画のみを収集したものです。アノテーターは独自のGUI（グラフィカル・ユーザー・インターフェース）ツールを用いて、あらかじめ定義された11種類の親ラベルと子クラスから適切なアノテーションを選択しました。10秒以上の動画については、開始時刻を動画終了の10秒前に設定した。また、交通事故のない映像には、時刻、天候、車線のラベルのみを付与した。

%マルチタスク/ラベルの定義
\cRyo{
\subsection{Annotation Definitions}
We have defined 11 parent categories of environmental information for the collected videos containing traffic accidents, each with several child categories. This section provides a description of the parent categories in boldface type, and the child categories are described in the following sections.

\noindent\textbf{Traffic Incident:}
This category indicates whether or not an accident is included in the video, with two subcategories: positive and negative. Videos in which a traffic accident occurs are assigned the category ``positive," and videos in which no traffic accident occurs are assigned the category ``negative."

\noindent\textbf{Incident Targets:}
This parent category indicates the traffic accident captured by the dashcam at the time of the accident. For example, the subcategories ``Car" and ``Pedestrian" are included.

\noindent\textbf{Contact Level:}
This parent category describes the severity of the crash, with three child categories: ``wreck" for significant collisions, ``touch" for minor collisions, and ``near miss" for videos that avoid a traffic accident.

\noindent\textbf{Environment:}
This parent category indicates the location of the accident in the video, with six child categories such as ``Major road," ``Highway," and ``Parking lot."

\noindent\textbf{Derivation Object:}
This parent category indicates the viewpoint of the in-vehicle camera, with two child categories: ``My car vs Another" when the traffic accident video is recorded from the first-person perspective and ``Another vs Other" when the video is recorded from the third-person perspective (i.e., someone else's accident captured by the dashcam of your car).

\noindent\textbf{Predictability:}
This parent category indicates whether the annotator can predict traffic accidents, with ``Predictable" and ``Not predictable" as child categories. ``Predictable" indicates that the annotator could have predicted the traffic accident in video two to three seconds before it occurred. If the annotator could not have guessed the traffic accident, ``Not Predictable" is assigned.

\noindent\textbf{Reaction:}
This parent category evaluates the car's behavior recorded in the traffic accident video, with three child categories: ``Avoidance" when a car avoids a traffic accident after it has occurred, ``stoppage" when the car stops before the accident occurs, and ``cancellation" when the car decelerates.

\noindent\textbf{State:}
This parent category evaluates the cause of the accident, with eight child categories such as ``ignored traffic light," ``cutting off," ``sudden stop," ``inattention," ``overspeed," and ``skid."

\noindent\textbf{Time:}
This parent category indicates when the traffic accident occurred, with two child categories: ``daytime" and ``night."

\noindent\textbf{Traffic Lane:}
This parent category represents the lane where the traffic accident was filmed, with two child categories: ``left" and ``right."

\noindent\textbf{Weather:}
This parent category describes the climatic conditions when the traffic accident video was shot, with four child categories: ``sunny/cloudy," ``snowy," ``foggy," and ``undefined" for videos that do not include traffic accidents.

Of the parent classes defined above, ``Environment," ``Time," ``Traffic Lane," and ``Weather" are also annotated for videos that do not include traffic accidents. Child categories that the annotator could not label are annotated as ``Undefined."
}

% \begin{figure*}[t]
% \centering
% \includegraphics[width=0.98\linewidth]{figure/each_labels_acc.png}
% \caption{Performance evaluation of environmental labels. By comparing the accuracy levels and the classification of positive/negative and environmental labels, we could identify labels that are effective in improving classification accuracy and, thus, our evaluations.
% % The experimental results show that eight of the 10 labels we provided improved the positive/negative classification performance.
% }
%  \label{fig:exp2}
% \end{figure*}

\subsection{Dataset Statistics}

\cNishiyama{This section presents the statistics of the proposed V-TIDB. Details of each class are shown in Figure \ref{fig:statistic_vtidb}. The V-TIDB consists of 9062 videos, with 4088 showing traffic accidents (positive class) and 4974 showing no traffic accidents (negative class). The dataset has a slightly larger negative class video. The parent category ``Incident Target" has the most significant number of classes, followed by ``State."}

% ここ上の表の引用がおかしくなってしまう
\cRyo{
The parent category ``Incident target" contains ten child categories, with ``car" being the most common, accounting for about 77\% of the total. ``Pedestrian" is the next most common, accounting for approximately 15\% of the total. ``Bicycles," ``animals," ``ambulances," and ``police cars" are included but appear rarely.

The parent category ``Contact level" consists of three child categories, with ``Wreck" being the most common, accounting for about 46\% of the total. ``Near-Miss" and ``Touch" account for about 27\% and 22\% of the total, respectively.

The parent category ``Environment" has six classes and is also assigned to images with no accidents. ``Major road" is the most frequent, accounting for about 65\% of the total, followed by ``Highway" accounting for approximately 32\% of the total. ``Parking lot," ``Gravel road," and ``Other location" appear rarely and account for less than 1\% of the total.

The parent category ``Derivation Object" has two child categories, with ``My car vs Another" accounting for 51\% of the total, and ``Another vs Others" accounting for approximately 47\% from the first and third-person perspectives, respectively.

The parent category ``Predictability" has two child categories, with "Not Predictable" accounting for about 64\% of the total and ``Predictable" accounting for about 34\% of the total.

The parent category ``Reaction" consists of four child categories, with ``Stoppage" being the most common, accounting for about 53\% of the total, followed by ``Deceleration" accounting for about 17\% of the total. ``Other" and ``Avoidance" account for approximately 14\% and 13\% of the responses, respectively.

The parent category ``State" consists of eight child categories, with ``Inattention" being the most frequent, accounting for approximately 25\% of the total, followed by ``Cutting off" accounting for about 23\% of the total. ``Other causes," ``Slip," ``Blindspot," ``Ignore traffic light," and ``Sudden stop" account for the remaining categories.

The parent category ``Time" has two child categories assigned to videos with no traffic accidents. ``Daytime" accounts for about 70\% of the total, while ``Night" accounts for approximately 29\% of the total.

The parent category ``Traffic lane" is a label given to videos in which no traffic accidents occur and has two child categories: ``Right-hand traffic" accounts for about 59\% of the videos, while ``Left-hand traffic" accounts for approximately 38\%.

The parent category ``Weather" has four child categories. ``Sunny/Cloudy" is the most common, accounting for about 87\% of the total. ``Rainy" is the next most common, accounting for approximately 9\%, followed by ``Snowy," which accounts for around 3\% of the total. ``Foggy" accounts for about 0.1\%.

The V-TIDB statistics show a significant bias in the child categories. However, we created a large multi-label dataset with environmental information to determine the causes of traffic accidents.
}
% In the next section, we analyze the recognition results of actual traffic accident videos utilizing the environmental information assigned as multi-labels and provide benchmarks. Our analysis of traffic accident recognition aims to elucidate the causes of traffic accidents.

% \begin{figure*}[t]
% \centering
% \includegraphics[width=0.98\linewidth]{figure/Accuracy03.png} 
% \caption{Accuracy evaluation of each class. The accuracy of positive and negative classifications for each class was compared by evaluating their accuracy.
% % In this experiment, we showed that the 'animal', 'ignore traffic light', 'blindspot', and 'foggy' classes had high accuracy. In contrast, 'forest' and 'sudden stop' classes showed low accuracy.}
% }
% \label{fig:exp3}
% \end{figure*}

\begin{table}[t]
\centering
\caption{\cNishiyama{F1 score for predicting traffic accidents (positive/negative) by ResNet18 and ResNet50 without (baseline) and the addition of environmental information (marked with $\lozenge$).}}
\vspace{-10pt}
\begin{tabular}{lcc} \toprule
\label{tab:recognition_traffic_incident}

 & F1-score(ResNet18) & F1-score(ResNet50) \\ \midrule
                 
Traffic incident (Baseline)          & 0.969   & 0.912   \\
$\lozenge$ Incident target   & 0.963 (\color{blue}-0.006)   & 0.952 (\color{red}+0.040)  \\
$\lozenge$ Contact level     & 0.960 (\color{blue}-0.009)  & 0.956  (\color{red}+0.044) \\
$\lozenge$ Environment       & 0.971 (\color{red}+0.002)  & 0.940  (\color{red}+0.028) \\
$\lozenge$ Derivation object & 0.965 (\color{blue}-0.004)  & 0.954  (\color{red}+0.042)  \\
$\lozenge$ Predictability    & 0.977 (\color{red}+0.008)  & 0.968  (\color{red}+0.056) \\
$\lozenge$ Reaction          & 0.972 (\color{red}+0.003)  & 0.967  (\color{red}+0.055) \\
$\lozenge$ State             & 0.974 (\color{red}+0.005)  & 0.967  (\color{red}+0.055) \\
$\lozenge$ Time              & 0.970 (\color{red}+0.001)  & 0.889  (\color{blue}-0.023) \\
$\lozenge$ Traffic lane      & 0.957 (\color{blue}-0.012)  & 0.945  (\color{red}+0.033) \\
$\lozenge$ Weather           & 0.971 (\color{red}+0.002)  & 0.922  (\color{red}+0.010)  \\ \bottomrule

\end{tabular}
\vspace{-10pt}
\end{table}

\section{Experiments}

%In this section, we describe experiments to evaluate the performance of our proposed V-TIDB for traffic accident recognition. In order to clarify which labels contribute to improving the accuracy of traffic accident prediction, we compare the results of learning with only positive and negative labels and the results of learning with labels for each type of environmental information.

\cNishiyama{
%本章では，我々の提案する交通事故認識用データセットV-TIDBの性能を評価する実験について説明する．V-TIDBに付与したどの環境情報ラベルが交通事故認識の精度向上に寄与するかを明らかにするための比較実験を行う．比較実験は``Incident target"と``Contact level"について行う．``Incident target",``Contact level"のみで学習した結果とこれらのラベルに加えてV-TIDBに付与したマルチラベルとして付与した環境情報をマルチタスク学習した結果を比較する．
In this chapter, we describe an experiment to evaluate the performance of our proposed V-TIDB dataset for traffic accident recognition and conduct a comparison experiment to determine which environmental information labels assigned to V-TIDB contribute to improving the accuracy of traffic accident recognition. Comparison experiments are conducted for ``Incident target" and ``Contact level." We compare the results of multi-task learning of environmental information added as multi-labels to V-TIDB with the results of learning only ``Incident target" and ``Contact level."
}

\subsection{Implementation details}
\cNishiyama{
We used 3D-ResNet\cite{3DResnet}, a simple yet powerful framework for video action recognition, to recognize traffic accidents in videos. We added a linear layer of the number of parent categories of the environmental information added to the final layer of 3D-ResNet for multi-task learning of the multi-label environmental information assigned to V-TIDB, and the output features of the linear layer are trained as the number of child categories. The network was given a video size of $112 \times 112$ pixels and 16 frames as input. The optimizer used stochastic gradient descent (SGD) with learning rate and momentum set to 0.01 and 0.9, respectively. Cross-entropy was used as the loss function. When the network was trained with multiple labels, the loss function was defined as the average value for the loss value of each label. The batch size was set to 128. The evaluation metric was the F1-score, which in this paper is defined as the average output and label accuracy of the network across 16 clips in the time direction.
}

\begin{table}[t]
\centering
\caption{\cNishiyama{Comparison between each parent category by f1 score of contact level recognition by ResNet18 between no environmental information (baseline) and the addition of environmental information (marked with $\lozenge$).}}
\vspace{-5pt}
\begin{tabular}{lccc} \toprule
\label{tab:recognition_contact_level}
                             & wreck & touch & near-miss \\ \midrule
Contact level(Baseline)      &0.55   & 0.15  & 0.51     \\
$\lozenge$ Incident target   &0.63 (\color{red}+0.08) &0.24 (\color{red}+0.09)&0.41 (\color{blue}-0.10)\\
$\lozenge$ Traffic incident  &0.60 (\color{red}+0.05)&0.30 (\color{red}+0.15)&0.44 (\color{blue}-0.07)\\
$\lozenge$ Environment       &0.60 (\color{red}+0.05)&0.16 (\color{red}+0.01)&0.49 (\color{blue}-0.02)\\
$\lozenge$ Derivation object &0.64 (\color{red}+0.09)&0.21 (\color{red}+0.06)&0.46 (\color{blue}-0.05)\\
$\lozenge$ Predictability    &0.63 (\color{red}+0.08)&0.17 (\color{red}+0.02)&0.27 (\color{blue}-0.24)\\
$\lozenge$ Reaction          &0.63 (\color{red}+0.08)&0.29 (\color{red}+0.14)&0.45 (\color{blue}-0.06)\\
$\lozenge$ State             &0.62 (\color{red}+0.07)&0.18 (\color{red}+0.03)&0.47 (\color{blue}-0.04)\\
$\lozenge$ Time              &0.55 (\color{red}+0.00)&0.33 (\color{red}+0.18)&0.36 (\color{blue}-0.15)\\
$\lozenge$ Traffic lane      &0.61 (\color{red}+0.06)&0.22 (\color{red}+0.07)&0.32 (\color{blue}-0.19)\\
$\lozenge$ Weather           &0.58 (\color{red}+0.03)&0.27 (\color{red}+0.12)&0.43 (\color{blue}-0.08)\\
\bottomrule

\end{tabular}
\vspace{-10pt}
\end{table}

\cNishiyama{
\subsection{\cRyo{Verification of the Effectiveness of Environmental Labels in Classifying the Presence or Absence of Traffic Accidents}}
We perform a comparison experiment to demonstrate the usefulness of multi-labeled detailed environmental information in traffic accident recognition. 

%表\ref{tab:recognition_traffic_incident}は``Incident target"(positive/negative)をベースラインとして括弧内にベースラインとの差分を表記した．差分を確認するとResnet18, 50の両モデルで複数の環境ラベルで精度向上が見られる.Resnet18では，10種類6種類がで精度向上が見られた．Resnet50では，``Time"を除く全てのラベルで性能が向上している． 具体的な環境ラベルとしては，``Environment", ``Predictability", ``Reaction", ``State", ``Time", ``Weather"で精度改善が見られた．このことから複数の環境ラベルの付与は交通事故認識において有用である．また，最も交通事故認識精度を向上させた親カテゴリは，``Predictability"であり，Resnet18では，0.008ポイントの向上がみられた．またResNet50も同様に"Predictability"が最も精度向上に寄与しており，0.056ポイントの向上が見られた．

\begin{figure*}[t]
\centering
\scalebox{1.0}{
% \hspace{-0.7cm}
\includegraphics[width=1.0\linewidth]{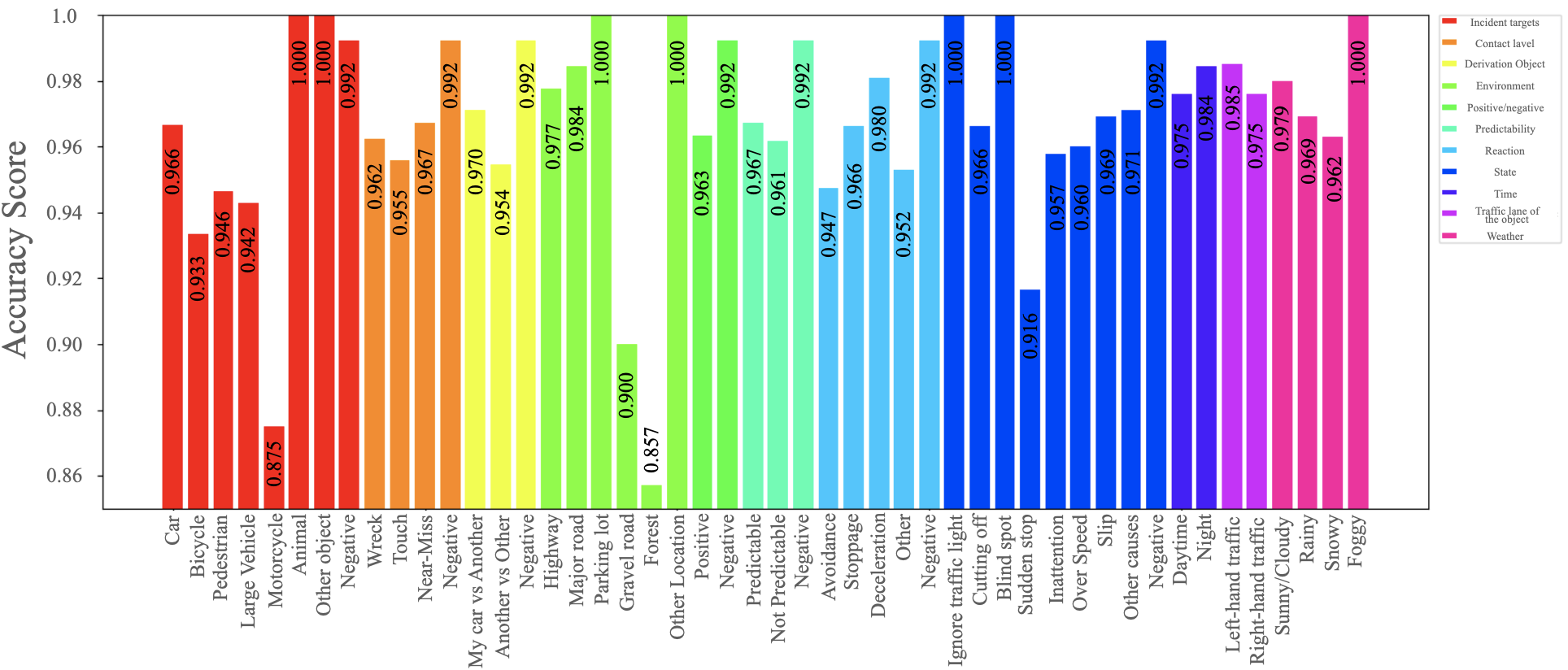} 
}
\caption{Accuracy evaluation of each class. The accuracy of positive and negative classifications for each class were compared by evaluating the their accuracy.
% In this experiment, we showed that the 'animal', 'ignore traffic light', 'blindspot', and 'foggy' classes had high accuracy. In contrast, 'forest' and 'sudden stop' classes showed low accuracy.}
}

\label{fig:exp3}
\vspace{-15pt}
\end{figure*}

%Table \ref{tab:recognition_traffic_incident} shows the difference between the baseline and the ``incident target"(positive/negative) in parentheses.The differences show that both 3D-ResNet18 and 3D-ResNet50 improved the accuracy of several environmental labels.3D-ResNet18 shows an increase in accuracy for 6 of the ten environmental labels, while 3D-ResNet50 shows an increase in performance for all labels except for ``Time". 

Table \ref{tab:recognition_traffic_incident} compares the F1-score of ``traffic incident" alone with the F1-score of V-TIDB plus the environmental label. The models used in the comparison experiment are 3D-ResNet18 and 3D-ResNet50. The F1-score of the model trained with only the parent category ``Incident target" is used as the baseline. The environmental information assigned to V-TIDB and to ``Incident target" is shown as $\lozenge$.
The parentheses to the right of the environmental information score marked with $\lozenge$ indicate the difference from the baseline.The differences indicate that both 3D-ResNet and 3D-ResNet50 have improved the accuracy of some environmental labels.For 3D-ResNet18, the F1-score improves with the addition of 6 out of 10 parent categories of environmental information, while for 3D-ResNet50, the performance improves with the addition of all parent categories except for ``Time".

For the specific environmental labels, the accuracy was improved for ``Environment", ``Predictability," ``Reaction," ``State," ``Time," and ``Weather."
Therefore, multiple environment labels are helpful for traffic accident recognition.
The parent category that improved the accuracy of traffic accident recognition the most was ``Predictability," and ResNet18 improved by 0.008 points. Predictability" also contributed the most to the accuracy improvement in ResNet50, with an improvement of 0.056 points.
}

% \vspace{-5pt}
\subsection{\cRyo{Verification of the effectiveness of environmental labels in classifying traffic accident levels}}
\vspace{-5pt}

\cNishiyama{
%我々は交通事故認識で推定した``Traffic incident"(positive/negative)からさらに一歩踏み込んだ交通事故のレベルを示す``Contact level"についての認識精度の比較実験を行う．実験は``Contact level"のみのF1-scoreと環境ラベルを付与したF1-scoreを比較する．比較実験で使用するモデルは``Traffic incident"(positive/negative)に対する交通事故認識精度の高いResNet18を選択した．ここで　``Contact level"の親カテゴリのみで学習した場合のF1-scoreを　Baselineとする．また，``Contact level"に追加したV-TIDBに付与されている環境情報を$\lozenge$によって示す．表\ref{tab:recognition_contact_level}から``wreck"では10種類中10種類の環境情報ラベルを付与した際に精度向上が見られ，最も精度向上に寄与した環境情報の親カテゴリは``Derivation object"であり0.09ポイントの向上が見られる．``touch"では，10種類中10種類の環境情報ラベルを付与した際に精度向上が見られ，最も精度向上に寄与した環境情報の親カテゴリは，``Traffic incident"(positive/negative)であり，0.15ポイント向上した．``near-miss"は環境情報を付与したことによって認識精度は下がる結果となった．とくに精度を下げている環境情報の親カテゴリは，``Predictability"であった．
We will experiment to compare the recognition accuracy of ``Contact level", which indicates the level of a traffic incident, with that of ``Traffic incident" (positive/negative) estimated by the traffic incident recognition. The experiment compares the F1-score of ``Contact level" alone with the F1-score with environmental labels. The model used in the comparison experiment is 3D-ResNet18, which has high recognition accuracy for ``traffic incident" (positive/negative).
The F1-score of the model trained with only the parent category of ``Contact level" is defined as the baseline. The environmental information assigned to V-TIDB added to ``Contact level" is shown by $\lozenge$. Table \ref{tab:recognition_contact_level} shows that the accuracy of ``wreck" improved when ten environmental information labels out of ten were added. The parent category of environmental information that contributed the most to the improvement in accuracy was the ``Derivation object,'' with an improvement of 0.09 points. In the ``touch'' category, accuracy was improved when ten out of ten environmental information labels were assigned. The parent category of environmental information that contributed the most to accuracy was ``traffic incident'' (positive/negative), with an improvement of 0.15 points. The recognition accuracy of ``near-miss" was decreased by adding the environmental information. The parent category of the environmental information that reduced the accuracy was ``predictability."
}

\subsection{Analysis of incident recognition performance on each environment}

In this section, we discuss the different classification accuracies of the environmental labels assigned to V-TIDB as multi-labels. Accuracy for ``traffic incident" was calculated for each child category of the environmental labels. Figure ~\ref{fig:exp3}  shows the child categories on the horizontal axis. Each parent category is color-coded. The vertical axis is the value of Accuracy. The network is trained for traffic accident recognition.
The classes with the highest accuracy in classifying traffic accidents were ``animal", ``ignored traffic light", ``blindspot", and ``foggy", which showed correct classification results for all validation data. In contrast, the ``motorcycle", ``forest", and ``sudden stop" classes showed low accuracy. The ``Negative" child category of each label indicates a video in which no traffic accidents have occurred.

% ここでは、各環境ラベルの精度の違いについて説明する。
% 
% このため、環境クラスごとに正負の分類精度を算出した。Fig.~ref{fig:exp3} は交通事故分類の各クラスに対する精度スコアをまとめたものである。なお、ネットワークは交通事故分類のためにのみ学習されている。交通事故分類の精度が最も高いクラスは、「動物」、「信号無視」、「死角」、「霧」で、全ての検証データで正しい分類結果を示しています。一方、「バイク」、「森林」、「急停車」クラスは低い精度を示した。各ラベルの否定クラスは、アノテーターがラベリングに困難を感じたものを指す。ただし、検証データセットのクラス数が同じであったため、同じ値をとった。

\section{Discussion}
\textbf{Effectiveness of Environmental Information Labels in Traffic Accident Recognition}
\cNishiyama{
% 本論文の強みについて言及する
%``Traffic incident"をベースラインとした一つ目の実験では，環境ラベルを追加することで精度向上を見られる親カテゴリはResNet18において10個中6個存在した．また，ResNet50では，10個中9個の存在した．このことから，交通事故認識においてV-TIDBの周囲の詳細な環境情報やアノテーターの主観を含むアノテーションは交通事故認識に対して有効である．特に交通事故認識精度の向上に寄与した``Predictability"はアノテーターの主観によって付与される2から3秒前の交通事故予測に基づいた親カテゴリである． これは表\ref{tab:compare-dataset}より他のデータセットには存在しない，我々の提案するデータセット特有のラベルであり主観性を含めたアノテーションが交通事故認識において有用であることを示している．

%``Contact level"をベースラインとした２つ目の実験である交通事故のレベルの認識では，環境情報を追加することで``wreck"や``touch"などの交通事故の規模の認識において有効に働いた．一方で交通事故が起こる直前に回避したことを示す``near-miss"の認識には悪影響を及ぼしていた．
% ``near-miss"において認識精度を下げる原因として考えられるのは，ロバストな``near-miss"の認識において環境情報がそれほど重要でないことが示唆される．交通事故の認識の場合は，環境を要因(例えば，雨が降っているから事故が起きた．夜で視界が狭いから事故が起きた)とする事故動画に対する認識性能を向上が期待できる．一方で，``near-miss"については，運転者の注意不足による可能性が多く，そういった動画の認識は，環境に依存せず，むしろ環境に依存するように学習することが認識において悪影響であったため，全体的に精度が低下する結果となったと考える．
% また，交通事故の発生の有無の認識だけではなく，交通事故のレベルの認識は精度向上の余地がある．交通事故のレベルの認識率を上げる方法として，マルチラベルとして付与した詳細な環境情報の組み合わせや，すべてのマルチラベルを使用して学習する方法を考えている．

In the first experiment with ``traffic incident" as the baseline, there were six out of ten parent categories in 3D-ResNet18 for which adding an environmental label improved accuracy. In 3D-ResNet50, there were nine out of ten parent categories. This indicates that annotations including detailed environmental information around V-TIDB and the annotator's subjective view are effective for traffic accident recognition. In particular, ``Predictability'', which contributed to the improvement of traffic accident recognition accuracy, is a parent category based on the annotator's subjective prediction of a traffic accident two to three seconds before. This is a unique label for our dataset, which does not exist in any other dataset than table \ref{tab:compare-dataset}, indicating that the annotations including subjectivity are useful for traffic accident recognition.

In the second experiment, using ``Contact level" as a baseline, the addition of environmental information was effective in recognizing the scale of traffic accidents such as ``wreck" and ``touch". On the other hand, it had a negative effect on the recognition of ``near-misses," which indicate that the driver avoided a traffic accident just before it occurred. 
\cRyo{One possible cause for the decrease in recognition accuracy in "near-miss" is suggested to be that environmental information is not as important for robust recognition of "near-miss," and may even be a hindrance. In the case of traffic accident recognition, improving recognition performance for accident videos that attribute environmental factors (such as "the accident happened because it was raining" or "the accident happened because it was dark and visibility was poor") is expected. On the other hand, for "near-miss," there are many possibilities of driver's inattention, and recognizing such videos by learning to depend on the environment had a negative impact on recognition accuracy, resulting in an overall decrease.}
In addition to the recognition of whether or not a traffic accident has occurred, there is room for improving the accuracy of the recognition of the level of traffic accidents.
As a method to increase the recognition rate of the level of traffic accidents, we are considering a combination of detailed environmental information assigned as multi-labels or a method of learning using all multi-labels.

}

\cNishiyama{
\textbf{Comparison issues between traffic accident recognition datasets.}
% 交通事故認識データセット間の比較の問題
We have considered comparative experiments on several datasets for traffic accident recognition. However, some datasets were ``not downloadable'' because they were difficult to compare.
Some datasets were ``downloadable but corrupted''. In addition, some datasets were ``private''.
Comparative experiments were complex in our environment for the above reasons.

The contribution of this paper is not to propose a conventional dataset to mark the traffic accident recognition rate but rather
1) It reveals that various environmental labels are effective in traffic accident recognition.
2) We collect more dashcam videos and IDs than a conventional dataset, label them with the presence of traffic accidents and environmental information, and disclose this information to the public.
}

%我々は，交通事故認識を行うために，複数のデータセットでの比較実験を検討しました．しかし，比較が困難であった理由として，``ダウンロードが不可能"なデータセットが存在した．また，``ダウンロード可能だが破損している"データセットも存在した．加えて``非公開"のデータセットも存在した．そのため我々の環境では，比較が困難だった．

% 本論文の貢献は，従来の交通事故認識率をマークするデータセットの提案ではなく，
% 1) 交通事故認識において多様な環境ラベルが効果的であるという事を明らかにした 2) 従来のデータセット以上のダッシュカム動画とIDを収集し，それらに交通事故の有無及び環境情報のラベルリングを行い，それらの情報公開することに注意していただきたい．

\section{CONCLUSION}
\cNishiyama{
%本研究では、交通事故認識の精度を向上させるために、9000以上の映像からなる大規模データセットであるV-TIDBを提案した。本研究では交通事故そのものを示す親カテゴリである``Traffic incident"と交通事故の規模を表す親カテゴリである``Contact level"についてV-TIDBで付与したマルチラベルな環境情報を追加したf1-scoreによる比較実験をおこなった．比較実験の結果``Traffic incident"では10種類のうち6種類以上の環境情報で交通事故の認識精度が向上した．また``Contact level"の``wreck"と``touch"では，10種類中10種類の環境情報で交通事故の規模認識精度が向上した．我々の提案するV-TIDBは今後公開予定である．
In this study, we proposed V-TIDB, a large-scale dataset consisting of more than 9000 videos, to improve the accuracy of traffic accident recognition. In this study, a comparison experiment using the f1-score was conducted for ``Traffic incident,'' which is the parent category of traffic accidents themselves, and ``Contact level,'' which is the parent category of the scale of traffic accidents, by adding multi-label environmental information assigned by V-TIDB. The results showed that the accuracy of traffic incident recognition was improved for more than six out of ten types of environmental information in the ``Traffic incident'' category. In addition, for ``contact level'' of ``wreck'' and ``touch'', ten out of ten types of environmental information improved the recognition accuracy of the traffic incident scale.
% Our proposed V-TIDB will be released soon.
}
% \addtolength{\textheight}{-12cm}
% This command serves to balance the column lengths
 % on the last page of the document manually. It shortens
 % the textheight of the last page by a suitable amount.
 % This command does not take effect until the next page
 % so it should come on the page before the last. Make
 % sure that you do not shorten the textheight too much.

\section*{Acknowledgement}
Computational resource of AI Bridging Cloud Infrastruc- ture (ABCI) provided by National Institute of Advanced In- dustrial Science and Technology (AIST) was used.

%%%%%%%%%%%%%%%%%%%%%%%%%%%%%%%%%%%%%%%%%%%%%%%%%%%%%%%%%%%%%%%%%%%%%%%%%%%%%%%%

%%%%%%%%%%%%%%%%%%%%%%%%%%%%%%%%%%%%%%%%%%%%%%%%%%%%%%%%%%%%%%%%%%%%%%%%%%%%%%%%

% \begin{thebibliography}{99}
% \bibitem{DRIVE}Wentao Bao, Qi Yu, Yu Kong. : DRIVE: Deep Reinforced Accident Anticipation with Visual Explanation. In : ICCV(2021)

% \bibitem{SA} Chan, F.H., Chen, Y.T., Xiang, Y., Sun, M.: Anticipating accidents in dashcam videos. In: ACCV (2016)

% \bibitem{A3D}Yao, Y., Xu, M., Wang, Y., Crandall, D.J., Atkins, E.M.: Unsupervised traffic accident detection in first-person videos. In: IROS (2019)

% \bibitem{DADA}Fang, J., Yan, D., Qiao, J., Xue, J.: Dada: A large-scale benchmark and model for
% driver attention prediction in accidental scenarios. arXiv:1912.12148 (2019)

% \bibitem{CCD}Wentao Bao , Qi Yu , Yu Kong , : Uncertainty-based Traffic Accident Anticipation with Spatio-Temporal Relational Learning. arXiv:2008.00334v1 In ACM MM (2020)

% \bibitem{DoTA} Yao, Yu, et al. "When, where, and what? a new dataset for anomaly detection in driving videos." arXiv preprint arXiv:2004.03044 (2020).

% \bibitem{NIDB}Tomoyuki Suzuki, Hirokatsu Kataoka, Yoshimitsu Aoki, and Yutaka Satoh. : Anticipating Traffic Accidents with Adaptive Loss and Large-scale Incident DB. In: CVPR(2018)

\bibliographystyle{splncs04}
\bibliography{egbib}

% \end{thebibliography}

\end{document}